\documentclass[letterpaper, 10 pt, conference]{ieeeconf}  

\usepackage[T1]{fontenc}
\usepackage{xcolor}
\usepackage{graphics}
\usepackage{graphicx} 
\usepackage{rotating}
\usepackage{dcolumn}
\usepackage{longtable}
\usepackage{multirow}
\usepackage{amsmath}
\usepackage{amssymb}    
\usepackage{xspace}
\usepackage{textcase}
\usepackage{comment}
\usepackage{float}
\usepackage{tikz}
\usepackage{overpic}
\usetikzlibrary{trees}
\usetikzlibrary{arrows}
\usepackage{hyperref} 
\usepackage{caption}
\usepackage{subcaption}
 \usepackage{url}
\usepackage{listings}
\usepackage[ruled,vlined,linesnumbered,noresetcount]{algorithm2e}

\DeclareMathOperator{\NN}{NN}
\DeclareMathOperator{\ODESolve}{ODESolve}

\DeclareMathOperator{\DAESolve}{DAESolve}

\newtheorem{theorem}{Theorem}[section]

\newtheorem{assumptions}{Assumptions}

\SetKwInOut{Parameter}{Parameters}
\SetKwFunction{DAEFn}{DAESolve}
\SetKwFunction{ODEFn}{FwdEuler}
\SetKwFunction{ODESolveAlg}{ODESolve}
\SetKwFunction{concat}{Concat}
\SetKwFunction{criterion}{Criterion}

\IEEEoverridecommandlockouts                              
\title{\LARGE \bf
Learning Neural Differential Algebraic Equations via Operator Splitting 
}

\author{James Koch$^{1}$, Madelyn Shapiro$^{1}$, Himanshu Sharma$^{1}$, Draguna Vrabie$^{1}$, J\'an Drgo\v na$^{1,2}$
\thanks{$^{1}$Pacific Northwest National Laboratory,}
\thanks{$^{2}$Department of Civil and Systems Engineering, and the Ralph O’Connor Sustainable Energy Institute, Johns Hopkins University, \texttt{jdrgona1@jh.edu}}
}

\begin{document}

\maketitle

\begin{abstract}
    Differential algebraic equations (DAEs) describe the temporal evolution of systems that obey both differential and algebraic constraints. Of particular interest are systems that contain implicit relationships between their components, such as conservation laws.  Here, we present an Operator Splitting (OS) numerical integration scheme for learning unknown components of DAEs from time-series data.  In this work, we show that the proposed OS-based time-stepping scheme is suitable for relevant system-theoretic data-driven modeling tasks. Presented examples include (i) the inverse problem of tank-manifold dynamics and (ii) discrepancy modeling of a network of pumps, tanks, and pipes. Our experiments demonstrate the proposed method's robustness to noise and extrapolation ability to (i) learn the behaviors of the system components and their interaction physics and (ii) disambiguate between data trends and mechanistic relationships contained in the system.
\end{abstract}

\section{INTRODUCTION}
Modeling and simulation of systems via differential algebraic equations (DAEs) can be difficult, both in terms of formulation as well as numerical implementation. Common issues include (i) over- or under-constraining states and (ii) incorrect model specification, either of which can lead to a prohibitively challenging modeling task \cite{kunkel2006differential}. While many such algebraic relationships may be known (e.g., mass is conserved through a pipe), incorporating these algebraic relationships into well-established data-driven modeling paradigms is non-trivial. Here, measurement noise, partial observations, and missing or incorrect physics confound the ability to effectively build data-driven models that obey algebraic constraints. 
In practice, simplified models can be adopted that capture improperly specified or missing physics at a given fidelity. A common approach is to employ a closure model \cite{mellor1982development}, i.e., a surrogate for the full-fidelity physics inserted into a lower-fidelity model. Closure models have enabled major advancements in scientific computing but still traditionally rely upon rigorous reduction from first principles physics to derive required simplified relationships~\cite{bush2019recommendations}. 

\subsection{Contributions}
To address this issue, we present a novel framework for the data-driven modeling of DAEs. Our framework integrates ideas from neural timesteppers~\cite{liu2022hierarchical}, operator splitting methods~\cite{holden2010splitting}, and Physics-Informed Neural Networks (PINNs)~\cite{karniadakis2021physics} to accommodate algebraic relationships in the context of neural ordinary differential equations (NODEs)~\cite{chen2018neural}. We demonstrate the utility of these methods on two data-driven closure modeling tasks: (i) a tank-manifold problem, where one needs to `invert' the model to provide an estimate of a property of a model component, and (ii) a tank-pipe-pump network for learning unknown nonlinear interactions. Our specific contributions are:
\begin{itemize}
    \item Taking inspiration from the field of operator splitting, we extend neural timesteppers to integrate DAEs. We achieve that via a neural surrogate for algebraic constraints alongside an ODE solver for differential states.
    \item We demonstrate the method’s utility in data-driven parameter estimation tasks, including closure and inverse modeling of constrained network dynamics.
    \item We open-source the implementation of the proposed method~\footnote{\url{https://github.com/pnnl/NeuralDAEs}} to promote reproducibility and adoption.
\end{itemize}

The paper is organized as follows. Section \ref{sec:background} provides a brief background.
Section \ref{sec:problem} defines the problem of interest.
Section \ref{sec:methods} details our proposed methodology and its implementation. Example results are provided in Section \ref{sec:results}, followed by a discussion in Section \ref{sec:conclusion}.

\subsection{Related Work}
\subsubsection{Machine Learning for Dynamical Systems}
Methods for learning dynamics from data exist on a continuum from black-box models to white-box models~\cite{Legaard2023}. The black-box modeling task is performed without regard to the underlying system or domain knowledge and includes Recurrent Neural Networks (RNNs) and their variants (e.g., LSTMs) \cite{funahashi1993approximation}, neural state space models~\cite{Skomski2021,CHAKRABARTY20231490}, neural timesteppsers \cite{liu2022hierarchical}, Neural ODEs (NODEs)~\cite{chen2018neural,OLEARY2022111466,Suyong2021}, sparse symbolic regression \cite{brunton2016discovering}, and Koopman operator~\cite{AppliedKoopman2012}. Often, such models extract salient features in the training data to infer future states. However, satisfying algebraic constraints in these data-driven models is not trivial. 
White-box models use domain knowledge to constrain models to match expected dynamics, often leading to parameter-tuning problems \cite{ramsay2007parameter}. In gray-box models, recently referred to as physics-informed machine learning~\cite{Nghiem2023}, prior knowledge is incorporated in a manner to constrain a black-box model. 
Neural ODEs constructed with domain knowledge, termed \textit{Universal Differential Equations} \cite{rackauckas2020universal,koch2023structural}, are an attractive modeling platform for these problems because of their flexibility in encoding  structural priors.

\subsubsection{Data-driven Modeling of DAEs}
 If the DAE is well-characterized, the parameter estimation, inverse design, and control tasks are possible through an adjoint sensitivity analysis \cite{petzold2000sensitivity,cao2003adjoint,7068974}. This strategy is generally successful if (i) the index of the underlying DAE is low and (ii) all constituent physics are known. 
Recent works have begun to address some common issues regarding data-driven DAE-based models. Moya and Lin \cite{moya2023dae} use explicit Runge-Kutta time-stepping to combine semi-explicit DAEs into machine learning frameworks, which they demonstrate by modeling power networks. Xiao et al. \cite{xiao2022feasibility} construct neural ODE and DAE modules to forecast a regional power network. Huang et al. \cite{huang2023minn} adopt a simplified approach to the construction of a surrogate system for sequence-to-sequence mappings for DAEs. These studies share features of expressing the learning task in terms of a semi-explicit DAE with varying degrees of parameterization and integration of domain knowledge. In addition to these studies, Zhong et al. \cite{zhong2021extending} addressed contact modeling in multi-body dynamics, the authors in the work handled a constraint update architecture with similar explicit time-stepping as in Moya \cite{moya2023dae} and Xiao \cite{xiao2022feasibility}.

While the treatment of domain-specific problems differs, all rely on multi-step integration techniques separating differential and algebraic state updates. Recognizing this common structure, we recast these methods under operator splitting, where integration is decomposed into sequential sub-tasks. We propose a new class of neural timesteppers for data-driven modeling of DAEs with partially unknown dynamics.

\section{BACKGROUND}
\label{sec:background}
\subsection{Differential-Algebraic Equations}
Differential-Algebraic Equations (DAEs) are systems comprising differential and algebraic relationships that describe the evolution of a system's states. Specifically, we are concerned with semi-explicit DAEs written as:
\begin{equation} \label{eq:semi_explicit}
\begin{split}
    \frac{dx}{dt}  & = f\left( x, y, u \right), \\
    0 & = g\left(  x,y,u \right),
\end{split}
\end{equation}
where $x \in \mathbb{R}^{n_x}$ are the differential states of a system that evolve according to the vector field defined by $f: \mathbb{R}^{n_x}\times \mathbb{R}^{n_y}\times\mathbb{R}^{n_u} \to \mathbb{R}^{n_x} $, $y \in \mathbb{R}^{n_y}$ are the algebraic states, and $u \in \mathbb{R}^{n_u}$ are exogenous inputs. The algebraic states $y$ evolve such that the algebraic relationships, defined by $g(x,y,u) = 0$, are satisfied for all time $t$. 

A standard  DAE solution approach is to transform the system into a set of ODEs to be integrated with standard numerical solvers. This process is termed \textit{index reduction}~\cite{kunkel2006differential}. The index of a DAE is a distance measure between a DAE and its associated ODE: an index-1 DAE requires one differentiation of algebraic relationships to yield a consistent set of ODEs, an index-2 DAE requires two differentiations, etc. After the index reduction, the governing equations can be written as a set of first-order ODEs.

\subsection{Neural Differential Equations}
A Neural Ordinary Differential Equation (NODE) approximates the flow map of a dynamical system with a neural network architecture. Consider a dynamical system with states $x \in \mathbb{R}^{n_x}$. A NODE approximates the temporal dynamics of $x$ according to the ODE:
\begin{equation} \label{eq:node}
    \frac{d x}{dt} = f\left(x;\theta\right),
\end{equation}
where the parametric flow  $f: \mathbb{R}^{n_x} \rightarrow \mathbb{R}^{n_x}$ maps the states to their time derivatives with tunable parameters $\theta$. The solution to Eq. \eqref{eq:node} is the initial value problem:

\begin{equation}
    x^{(t+\Delta t)} = x^{(t)} + \int_{t}^{t+\Delta t} f\left(x;\theta\right) dt,
\end{equation}
where the superscript contained in parentheses denotes indexing by time.
This integration is performed with a standard numerical ODE solver, which we denote as $x^{(t+\Delta t)} = \ODESolve\left(f ,x^{(t)}; \theta\right)$.
 This is achieved through either (i) backpropagating residuals through the elementary functions of the explicit time-stepping integrator (reverse-mode autodiff) \cite{rudy2019deep,liu2022hierarchical}, (ii) an adjoint sensitivity definition and solution \cite{chen2018neural}, or (iii) forward-mode autodiff \cite{revels2016forward}.

\subsection{Operator Splitting Methods}
In operator splitting methods (OS), also called fractional step methods,  the evolution of a system is split into simpler, independent sub-tasks. The underlying ODE is then solved explicitly as the sequential solution of these sub-tasks, albeit at a decreased level of accuracy (i.e., splitting error \cite{perot1993analysis}). There are several well-established OS schemes. The simplest one, Lie–Trotter operator splitting \cite{holden2010splitting}, assumes that the vector field \( f \) can be decomposed into two operators:
\begin{equation} \label{eq:lie-trotter}
    \frac{dx}{dt} = A(x) + B(x),
\end{equation}
for the interval \( t \in [t, t + \Delta t] \). Where \( A \) and \( B \) are assumed to be locally Lipschitz continuous vector fields, ensuring the existence and uniqueness of solutions for each sub-step and the well-posedness of the splitting scheme.
The Lie–Trotter splitting scheme defines the sub-tasks as:
\begin{subequations}
\begin{align}
    \hat{x}^{(t+\Delta t)} &= \ODESolve\left( A(x), x^{(t)} \right), \\
    x^{(t + \Delta t)} &= \ODESolve\left( B(x), \hat{x}^{(t+\Delta t)} \right).
\end{align}
\end{subequations}
The first sub-task solves the ODE defined by the operator \( A(x) \) on the time interval \( [t, t + \Delta t] \). The second sub-task solves the ODE defined by the operator \( B(x) \) on the same time interval, but with initial conditions given by the result of the first sub-task. Many variations of this scheme exist for different applications and orders of accuracy. For example, Strang splitting \cite{strang1968construction} is a second-order scheme that uses half-steps (i.e., steps of \( \frac{\Delta t}{2} \)) in the above sequence.

\section{PROBLEM FORMULATION}
\label{sec:problem}
A fundamental assumption of a NODE~\eqref{eq:node} is that a direct relationship exists between each system state and its temporal derivative. This is a reasonable assumption for many cases, but poses challenges for partially-observed systems~\cite{dupont2019augmented} and application to DAEs. Index reduction is a reliable technique for recasting a DAE into a set of ODEs for standard numerical integration~\cite{mattsson1998physical}, but this is incompatible with the problem when the DAE governing equations are unknown or incomplete, leading to an unknown system of ODEs and required state augmentations.

To address this limitation, we aim to solve the following parameter estimation problem for neural DAEs (NDAE):
\begin{subequations}
    \begin{align}
    \label{eq:problem}
&\underset{\theta_f}{\text{min}}     && \int_{t_0}^{t_N} \Big( ||x - \hat{x}||_2^2   +||y - \hat{y}||_2^2   \Big) \\
&\text{s.t}    && 
    \frac{dx}{dt}  = f(x(t), y(t), u(t); \theta_f) , \label{eq:problem:ODE}\\
  &  &&  0  = g(x(t), y(t), u(t)), \\
  &  &&  x({t_0}) = x_0, \, \, \,  y({t_0}) = y_0,
  \label{eq:problem:algebra}
\end{align}
\end{subequations}
Where $\hat{x}, \hat{y}$ represent reference state variables generated by the unknown DAE system~\eqref{eq:semi_explicit}. The objective of the problem~\eqref{eq:problem} is to train the parameters $\theta_f$ such that the reference state trajectories in the time interval $[{t_0}, {t_N}]$ are closely approximated by the NDAE model given via Eq.~\eqref{eq:problem:ODE} and Eq.~\eqref{eq:problem:algebra}, respectively.

\begin{assumptions}
\label{assum:1}
Lets consider the unknown DAE system~\eqref{eq:semi_explicit} with \( f \) and \( g \) being continuously differentiable functions in a neighborhood of the point \( (x_0, y_0, u_0) \). Suppose the following conditions hold:
\begin{enumerate}
    \item {(Regularity)} The Jacobian matrix \( \frac{\partial g}{\partial y} \) is nonsingular at \( (x_0, y_0, u_0) \), i.e., $\det \big( \frac{\partial g}{\partial y}(x_0, y_0, u_0) \big) \neq 0 $.
    \item {(Consistency)} The initial values satisfy the algebraic constraint, i.e., \( g(x_0, y_0, u_0) = 0 \).
\end{enumerate}
\end{assumptions}

\begin{theorem}[Picard-Lindel\"of for index-1 DAEs~\cite{brenan1996numerical}]
\label{thm:PL:DAE}
Under the above assumptions, there exists a unique, continuously differentiable solution \( (x^{(t)}, y^{(t)}) \) defined on an interval \( [t_0, t_0 + \Delta t) \), for some \( \Delta t > 0 \), such that the DAE system is satisfied and \( (x^{(t_0)}, y^{(t_0)}) = (x_0, y_0) \). Furthermore, the algebraic constraint can be locally solved for \( y \) as a function of \( x \) and \( u \), thereby reducing DAEs into a system of ODEs $ \frac{\partial x}{\partial t} = f(x, h(x,u), u)$  to which the classical Picard-Lindel\"of Theorem~\cite{coddington1961ode} applies.
\end{theorem}

\textit{Remark.} Theorem~\ref{thm:PL:DAE} follows directly from the Implicit Function Theorem under Assumption~\ref{assum:1}, which guarantees that the DAE can be reduced to an index-1 form locally. We include this result for completeness and to formalize the assumptions under which the algebraic surrogate \( h \) is valid.


\section{METHOD}
\label{sec:methods}
In this paper, we propose an efficient operator splitting (OS) numerical integration scheme for the neural DAE parameter estimation problem given by Eq.~\eqref{eq:problem}.
Specifically,  we utilize the above Picard-Lindel\"of Theorem~\ref{thm:PL:DAE} to derive a novel operator splitting-based numerical integration method for data-driven modeling of index-1 DAEs.

\subsection{Operator Splitting for Differential Algebraic Equations}
Now consider the DAE system of Eq.~\eqref{eq:semi_explicit}. Similar to the approach of a Lie-Trotter splitting scheme, we split the evolution of a system's differential and algebraic states into sequential sub-tasks. The first sub-task updates the algebraic states using a trainable deep neural network (DNN). The second sub-task takes as input the updated algebraic states and performs a single-step integration. These two sequential sub-tasks define one complete forward pass; all system states are updated exactly once. 
More formally, consider a differentiable ODE solver, denoted as $\ODESolve$. Using operator splitting to solve the semi-explicit DAE in Eq.~\eqref{eq:semi_explicit}, we define the following neural time-stepper:
\begin{subequations} \label{eq:neural_dae}
    \begin{align}
        y^{(t+\Delta t)} & = h\left( x^{(t)},y^{(t)},u^{(t)} ; \theta_h\right), \\
        x^{(t+\Delta t)} &= \ODESolve\left(f ,\{x^{(t)},y^{(t+\Delta t)}, u^{(t)}\}; \theta_f\right),
    \end{align}
\end{subequations}
where \( h: \mathbb{R}^{n_x} \times \mathbb{R}^{n_y} \times \mathbb{R}^{n_u} \rightarrow \mathbb{R}^{n_y} \) is a neural surrogate for the algebraic update. In classical DAE theory, such a mapping exists locally under the Implicit Function Theorem, provided that \( \frac{\partial g}{\partial y} \) is nonsingular (see Assumption~\ref{assum:1}). This guarantees that, for fixed \( x \) and \( u \), the algebraic variable \( y \) can be implicitly expressed as a function \( h(x, u; \theta_h) \) satisfying \( g(x, h(x,u;\theta_h), u) = 0 \).
In our framework, we approximate this mapping $h$ using a neural network and train its parameters $\theta_h$ to minimize the residual \( \|g(x, y, u)\| \). This surrogate enables efficient algebraic updates in the first step of the operator splitting scheme. In contrast to PINNs~\cite{RAISSI2019686}, which minimize residuals for all equations simultaneously, our method separates the algebraic and differential components, handling the latter through the $\ODESolve$ step. This modular structure improves interpretability and computational flexibility when modeling partially known DAEs.

We denote this complete set of operations as $\DAESolve$, with one complete time step captured as:
\begin{equation}
    \begin{split}
        \{x^{(t+\Delta t)},y^{(t+\Delta t)}\} &= \\
        \DAESolve&\left(f,h,\{x^{(t)}, y^{(t)}, u^{(t)}\};\Theta\right),
    \end{split}
\end{equation}
 where $\Theta = \{\theta_h,\theta_f\}$ is the set of all model parameters, and the temporal integration bounds are assumed to be  $[t, t+ \Delta t]$.

\begin{figure}[ht]
\centering
        \begin{overpic}[width=1.0\columnwidth]{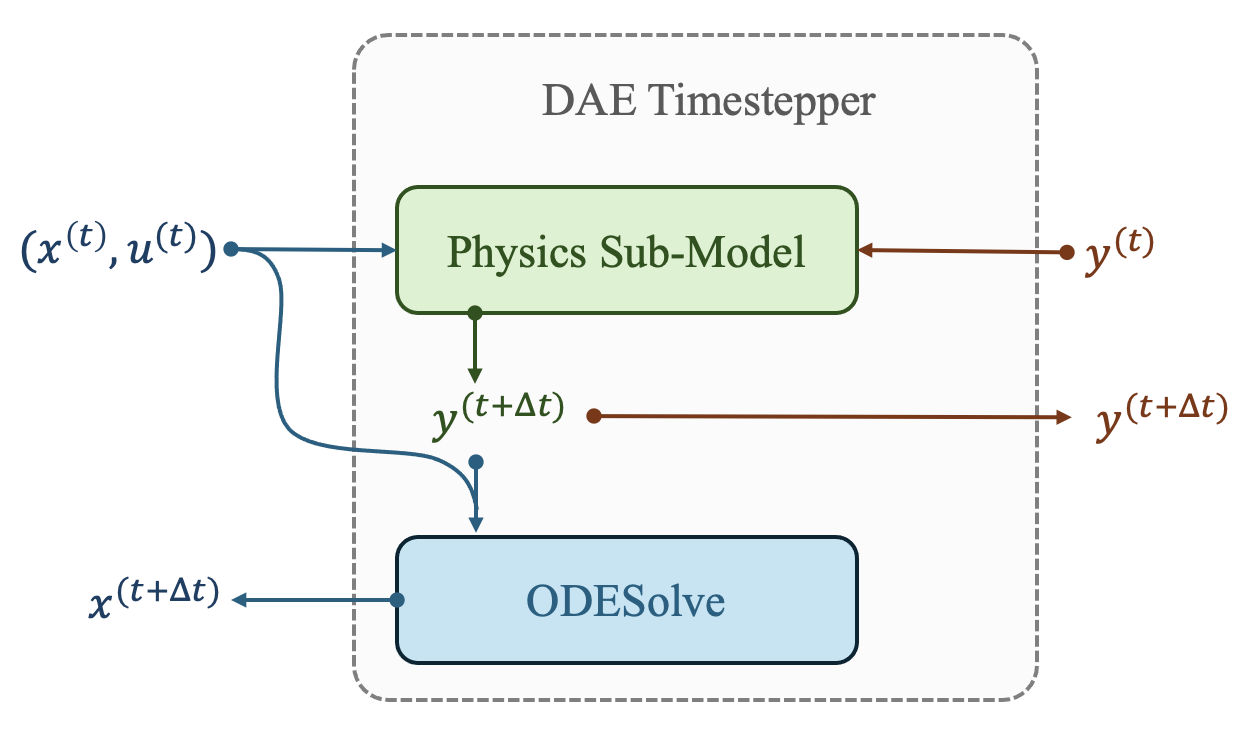}
        \end{overpic}  
        \caption{Schematic of the proposed operator splitting-based DAE Timestepper. }
        \label{fig:flow}
\end{figure}

Figure \ref{fig:flow} depicts the handling of differential and algebraic states through the proposed OS-based NDAE integration scheme. \
 In contrast, we adopt an operator splitting approach with learned algebraic surrogates and explicit time integration for DAE systems.
Note that in the proposed operator splitting scheme~\eqref{eq:neural_dae}, updates to the algebraic and differential states are performed sequentially: the algebraic state $y(t + \Delta t)$ is first updated via the algebra surrogate $h$, and the resulting value is then used to integrate the differential state $x(t + \Delta t)$ through the $\ODESolve$ step. This structure is reminiscent of Neural Controlled Differential Equations~\cite{kidger2020neural}, where external signals influence the dynamics via sequential updates. 
This ordering is non-commutative and plays a critical role in preserving the structure of the underlying DAE. The rationale stems from the formulation of semi-explicit DAEs~\eqref{eq:semi_explicit}, where the algebraic variable $y$ is not governed by a differential equation but is implicitly constrained to satisfy $g(x, y, u) = 0$ at all times. As such, the algebraic state must be updated first to ensure that the subsequent evaluation of the differential dynamics $f(x, y, u)$ begins from a consistent point on or near the constraint manifold.
Adopting this algebraic-first update reduces the constraint violations and aligns with standard practices in DAE solvers, which typically resolve algebraic updates before time-stepping~\cite{brenan1996numerical}.

Alternatives to the proposed scheme exist; most recently~\cite{pal2025} proposed a reverse-order update strategy in the form of a predictor-corrector scheme, where the differential state is first propagated using an unconstrained neural ODE model, and the result is then projected onto the constraint manifold to restore algebraic consistency. Alternatively, simultaneous updates as recently proposed by~\cite{lueg2025}, i.e., solving for $(x, y)$ jointly at each time step, reduce splitting error and provide tighter constraint satisfaction. However, this approach is more computationally demanding as it requires solving a constrained nonlinear optimization problem.

\subsection{Loss Function and Optimization Algorithm}
In this paper, solving the algebraic update sub-task is accomplished by formulating a PINN-style penalty loss~\cite{karniadakis2021physics} in which known constraints are used to `penalize' the discrepancy between constraint satisfaction and predicted states. These penalties are assessed in a multi-term loss function:
\begin{equation}
\label{eq:loss}
    \mathcal{L}(\Theta) = \lambda_1 \mathcal{L}_{\text{residual}}(X,\hat{X};\Theta) + \lambda_2 \mathcal{L}_{\text{constraints}}(X,\hat{X};\Theta),
\end{equation}
where $\mathcal{L}_{\text{residual}}(X,\hat{X};\Theta) = \sum_{t=0}^{N} \left( ||x_t - \hat{x}_t ||_2^2  + ||y_t - \hat{y}_t ||_2^2  \right)$ is a measure of the residual between the model's differential and algebraic states ${X} = \{({x}^{(0)},{y}^{(0)}), ..., ({x}^{(N)},{y}^{(N)})\}$, and the training dataset $\hat{X} = \{(\hat{x}^{(0)},\hat{y}^{(0)}), ..., (\hat{x}^{(N)},\hat{y}^{(N)})\}$,
where $N$ defines the length of a finite-horizon rollout.
The second term
$\mathcal{L}_{\text{constraints}}(X,\hat{X};\Theta) = \sum_{t=0}^{N} || g(x_t, y_t, u_t) ||^2_2$ is penalizing the constraint violations. The $\lambda$'s represent weightings for balancing the multi-objective loss function.

We optimize model parameters $\Theta$ with an off-the-shelf stochastic gradient-based optimizer~\cite{kingma2014adam} to solve $\Theta^* =  \underset{\Theta}{\text{argmin}} ~\mathcal{L}(\Theta)$.
For this work, we choose minimal architectures for ease of implementation and exposition, as outlined in Alg. \ref{alg:cap}. We use a first-order explicit timestepper for the ODE integration and use a DNN as a neural surrogate of an algebra solver. 
However, the proposed OS-based NDAE scheme is more generic than that and allows the combination of higher-order ODE solvers with iterative algebraic solvers.
\begin{algorithm}[htb!]
\caption{Operator splitting-based DAE training}\label{alg:cap}
\KwData{$\hat{X} = \{(\hat{x}^{(0)},\hat{y}^{(0)}), ..., (\hat{x}^{(N)},\hat{y}^{(N)})\}$: time series dataset with spacing $\Delta t$}
\Parameter{$\Theta = \{ \theta_f, \theta_h \}$: trainable parameters for mappings $f$ and $h$; \\ gradient step-size hyperparameter $\eta$; \\
loss function error tolerance $\epsilon$ }
\BlankLine
\SetKwProg{Fn}{Function}{:}{}
\tcc{Pick appropriate integrator}
\Fn{\ODESolveAlg{$f, \{ x, y, u\}$}}{
$x^{*}  \longleftarrow x + \Delta t \times f\left( x, y, u; \theta_f \right) $ \\
\textbf{return} $x^{*}$ 
}
\BlankLine
\tcc{Operator splitting via Eq.~\eqref{eq:neural_dae}}
\Fn{\DAEFn{$f, h, \{ x, y, u\}$}}{
$y^{*}  \longleftarrow h\left( x, y, u; \theta_h \right)$ \\
$x^{*}  \longleftarrow \ODESolveAlg\left( f, \{x, y^{*}, u\}; \theta_f \right)$ \\
\textbf{return} $(x^{*},y^{*})$ 
}
\BlankLine
\tcc{Training loop}
 \While{$\mathcal{L}(\Theta) > \epsilon$}{
    \tcc{Set initial conditions}
    $X \longleftarrow \{(\hat{x}^{(0)},\hat{y}^{(0)})\}$ \\
    \tcc{Forward pass}
    \For{k $\in 1, \dots, N$}{
        $X \longleftarrow $\concat($X$, \\ \DAEFn{$f, h, \{ x^{(k-1)}, y^{(k-1)}, u^{(k-1)}\}$} ) \\
}

\tcc{Evaluate loss via Eq.~\eqref{eq:loss}}
$\mathcal{L}(\Theta) \longleftarrow $ \criterion( $X,\hat{X}$ )\\

\eIf{$\mathcal{L}(\Theta) \le \epsilon$}{
break\;
}{

\tcc{Backward pass}
$ \nabla\mathcal{L}(\Theta) \longleftarrow \frac{\partial\mathcal{L}(\Theta)}{\partial \Theta} $

\tcc{Update model parameters}
$\Theta  \longleftarrow \Theta - \eta \nabla\mathcal{L}(\Theta)$
}
}
\end{algorithm} 

We acknowledge that the convergence of the residual loss \( \mathcal{L}(\theta) \) is not guaranteed in finite epochs, especially under noise or model mismatch. However, empirical studies suggest that with overparameterized networks and sufficient data, residual minimization can converge~\cite{wang2021understanding}, though adding data may not always improve performance due to inconsistency or overfitting. 
The neural surrogates \( h \) are trained to approximate solutions to the algebraic constraint in a region where the DAE remains index-1, as ensured by Assumption~\ref{assum:1}. 
While Theorem~\ref{thm:PL:DAE} assumes consistent initialization, in practice, we enforce approximate consistency by penalizing initial constraint violations via a residual loss. This encourages alignment with the constraint manifold over time, even under noise or partial observability.
Nevertheless, poorly trained surrogates may violate this regularity. Structural priors, as used in our case studies, or post-projection methods~\cite{pal2025}, can help ensure stable updates. 
These limitations motivate future work on robust residual weighting, adaptive stopping criteria, and formal convergence guarantees.

\section{NUMERICAL CASE STUDIES} \label{sec:results}
All experiments were performed using the NeuroMANCER Scientific Machine Learning (SciML) package~\cite{Neuromancer2023} built on top of PyTorch~\cite{paszke2019pytorch}. The open-source implementation of the presented method and associated numerical examples are available at GitHub\footnote{\url{https://github.com/pnnl/NeuralDAEs}}. Each presented model was trained using an Intel I9-9980 CPU laptop with a wall-clock training time on the order of 5 minutes.

\subsection{Tank-Manifold Property Inference}
\subsubsection{Problem Description}
\begin{figure}[h!]
\centering
        \begin{overpic}[width=0.7\columnwidth]{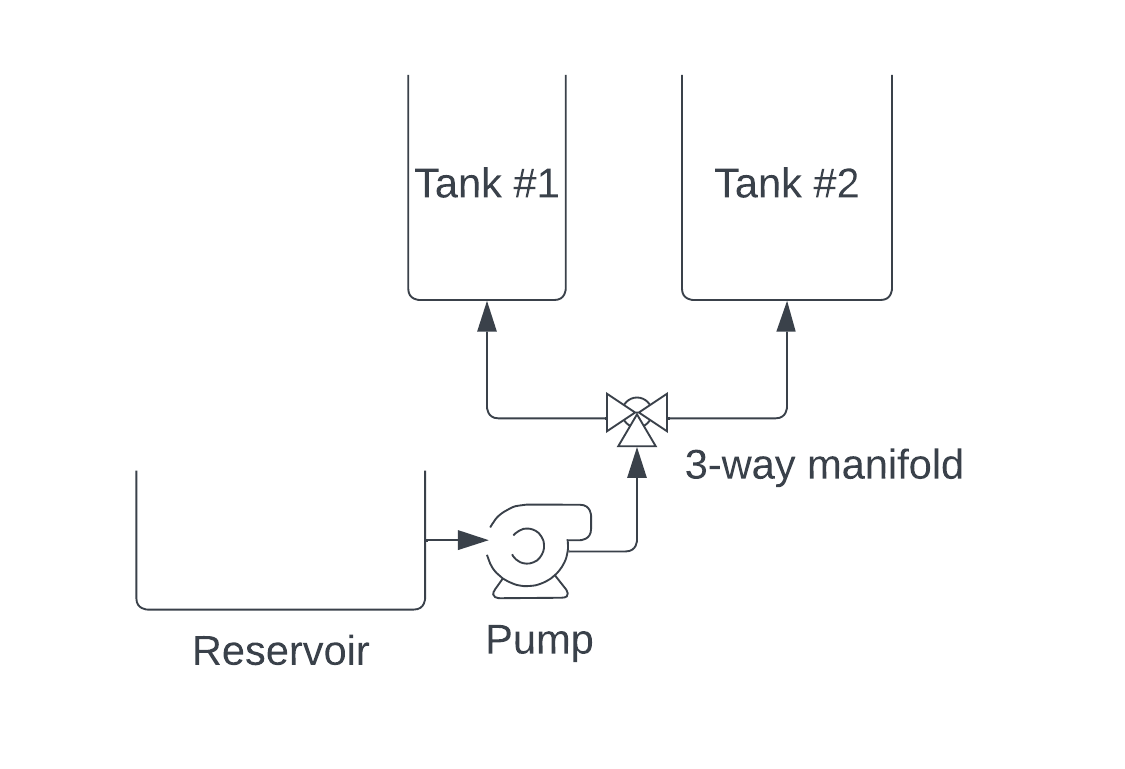}
        \end{overpic}  
        \vspace{-3mm}
        \caption{Tank-manifold-pump schematic. }
        \label{fig:manifold}
\end{figure}

A manifold is a physical realization of a conservation relationship that manifests as an algebraic constraint in a DAE. Figure \ref{fig:manifold} depicts a pair of tanks fed by a manifold. A pseudo-infinite reservoir supplies the pump. At the manifold, the inflow is exactly equal to the sum of the outflows; that is, $y_{\text{in}} = y_1 + y_2$, where $y$ represents volumetric flow rates and the subscripts $1$ and $2$ denote the two exit streams of the manifold. The two tanks have different area-height profiles but share a common datum such that the column heights are identical. As a fluid is pumped into the two tanks, the flow rates in the two streams vary to ensure (i) the conservation relationship of the manifold and (ii) identical pressure head of the tanks, i.e, $x_1 = x_2$, where $x$ is the height of the fluid. The full DAE that describes the evolution of this system is given as:
\begin{equation} \label{eq:manifold}
\begin{split}
    \frac{dx_1}{dt} = \frac{y_1}{\phi_1(x_1)}, \quad
    \frac{dx_2}{dt} = \frac{y_2}{\phi_2(x_2)},
\end{split}
\end{equation}
\vspace{-8mm}
\begin{align}%
0 = y_{\text{in}} - y_1 - y_2, \quad  0 = x_1 - x_2,  \nonumber
\end{align}%
 where $\phi_1$ and $\phi_2$ denote the area-height profiles for the two tanks. For this study, the area-height profiles are set to $\phi_1(x) = 3.0$ and $\phi_2(x) = \sqrt{x} + 0.1$ respectively. The goal of the learning task is to recover $\phi_2(x)$ given observations of tank heights $x$ and pump volumetric flow rates $y$. Note that if the task were to recover both $\phi_1(x)$ and $\phi_2(x)$ simultaneously, the problem would be ill-posed in the sense that there does not exist a unique set of tank profiles that recovers the behavior seen in observations.
\vspace{-3mm}

\subsubsection{Model Construction} \label{sec:model1}
In the system described by Eq. \eqref{eq:manifold}, the differential states are column heights $x_1$ and $x_2$, and the algebraic states are the volumetric flows $y_1$ and $y_2$. The input flow is treated as an exogenous control input, i.e., $u = y_{\text{in}}$. 
We aim to approximate the area-height profile of Tank \#1 as well as construct a surrogate for the behavior of the manifold such that (i) fluid is conserved and (ii) the fluid column height constraint is satisfied. There are two neural networks that we seek to train: $NN_1: \mathbb{R}^1 \rightarrow \mathbb{R}^1$, an approximation of the sought area-height profile $\phi_1$, and $NN_2: \mathbb{R}^4 \rightarrow (0,1)$, an approximation of the flow manifold connecting two tanks. Each is implemented as a feed-forward neural network with one hidden layer of 5 nodes with sigmoid activation functions. The output layer of $NN_1$ is linear while the output layer of $NN_2$ is sigmoidal, constraining the output to $(0,1)$.

\subsubsection{Optimization Problem}
A time series dataset $\hat{X} $ is constructed by simulation of the system given in Eq. \eqref{eq:manifold} regularly sampled on the interval $t\in[0, 500]$ with $\Delta t = 1.0$ and with a constant inflow of $y_{\text{in}} = u = 0.5$. The optimization problem seeks the minimum of the differential state trajectory reconstruction loss and algebraic state reconstruction loss when predicted by our $\DAESolve$:
\begin{equation} \label{eq:manifold_loss}
\begin{split}
\underset{\theta_1, \theta_2}{\text{min}} &~~ \lambda_1 \sum_{k=1}^{N}  ||x^{(k)} - \hat{x}^{(k)}||_2^2 + \lambda_2 \sum_{k=1}^{N}  ||y^{(k)} - \hat{y}^{(k)}||_2^2   \\
    \text{s.t.~~}    & \{x^{(k+1)},y^{(k+1)}\} \\ 
    ~ &=  \DAESolve\left(f,h,\{x^{(k)}, y^{(k)}, u^{(k)}\}; \{\theta_1, \theta_2 \}\right). \\
\end{split}
\end{equation}
The functions $f$ and $h$ are defined as:
\begin{equation}
\begin{split}
    f : 
\begin{cases}
\frac{y_1}{3}\\
\frac{y_2}{\NN_1(x_2;\theta_1)}\\
\end{cases},  \quad
    h : 
\begin{cases}
    u \NN_2(x_1,x_2,y_1,y_2;\theta_2) \\
    u \left(1 - \NN_2(x_1,x_2,y_1,y_2;\theta_2)\right) \\
\end{cases}. \\
\end{split}
\end{equation}
Note that the mapping $h$ defines a convex combination at the manifold: the outflows $y_1$ and $y_2$ will sum to $u$ by construction.
For all examples contained in this work, the optimizer is Adam~\cite{kingma2014adam} with a fixed learning rate of 0.001. Training is deemed `converged' after 20,000 epochs or until an early stop is triggered after 20 epochs of increasing loss. 

\subsubsection{Results}
\begin{figure}[htb]
    \centering
\includegraphics[scale=0.25,width=0.9\linewidth]{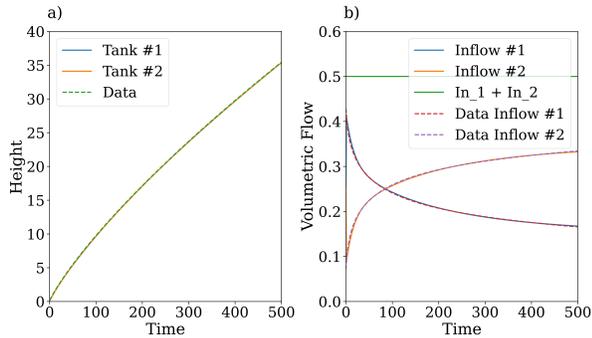}
    \caption{(a) Tank heights. (b) Volumetric flow.}
    \label{fig:manifold_heights_Flow}
\end{figure}

\begin{figure}[htb]
    \centering
\includegraphics[scale=0.25,width=0.9\linewidth]{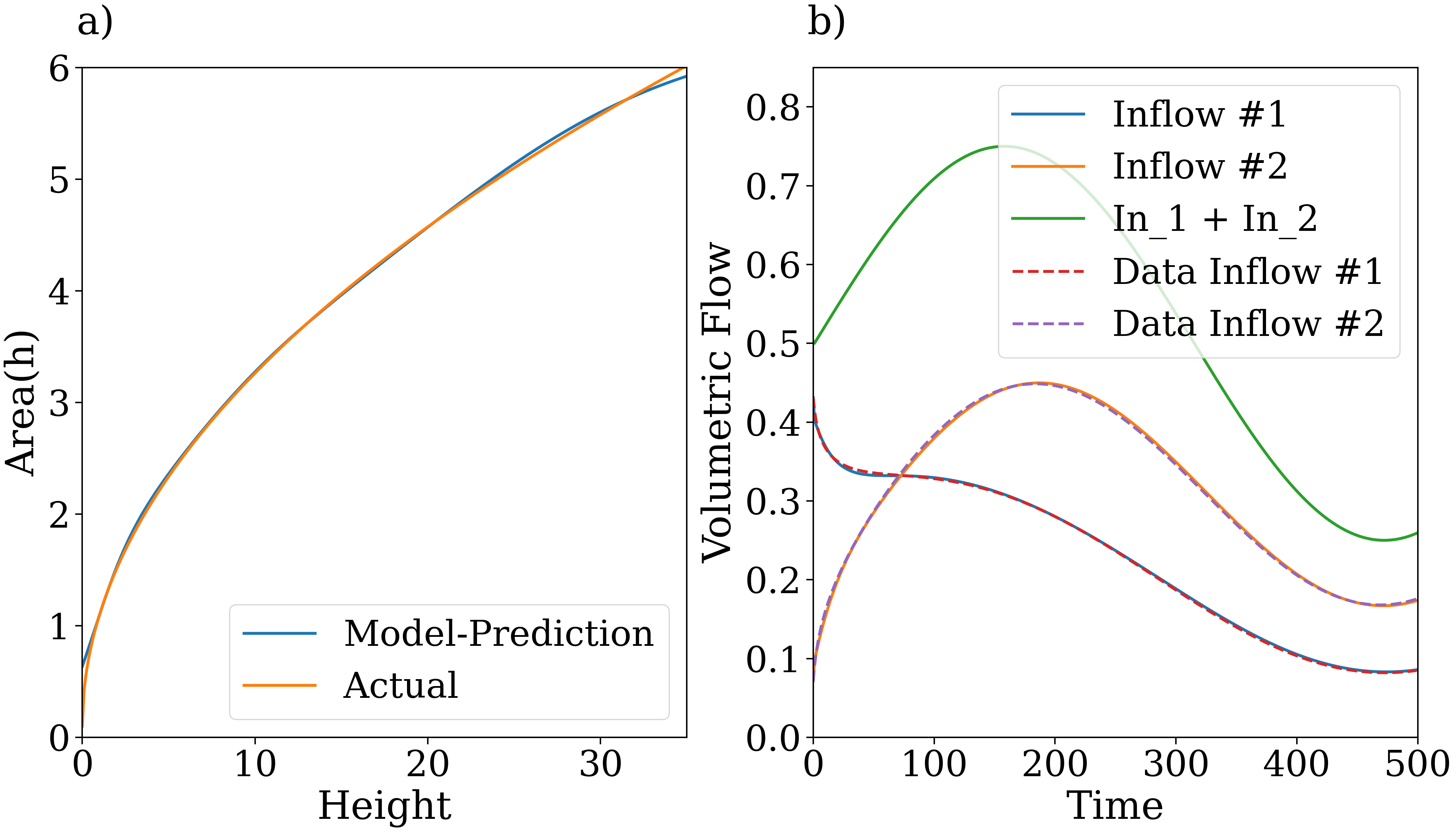}
    \caption{Trained model (a) Area-height relationships (b) Volumetric flow time series.}
    \label{fig:manifold_extrap_area_flow}
\end{figure}

The time series data for the tank heights and volumetric flows are given in Figs. \ref{fig:manifold_heights_Flow}(a) and \ref{fig:manifold_heights_Flow}(b) respectively.  The trained model accurately reproduces the time series data. Additionally, it is able to accurately learn the model of the unknown area-height profile of Tank \#2. Figure \ref{fig:manifold_extrap_area_flow}(a) depicts the ground-truth area-height profile and the model's reconstruction. Furthermore, to demonstrate the generalization, the trained model is tested on an unseen inlet mass flow rate; $u(t) = \frac{1}{2} + \frac{1}{4}\sin{\left( \frac{t}{100}\right)}$. The resulting time series of volumetric flows from the model rollout is compared with the ground-truth conditions in Fig. \ref{fig:manifold_extrap_area_flow}(b). Table-\ref{Table:MSE_tankManifold} gives the Mean Squared Error (MSE) of the trained neural DAE model.

Extrapolation to a new flow condition demonstrates the ability to (i) endogenize the behaviors of the components in the network and their interaction physics and (ii) disambiguate between data trends (such as correlations arising from specific operating regimes) and mechanistic relationships contained in the network.
\begin{table}[h]
\centering
\caption{Mean Square error for the differential state and algebraic states for the tanks-manifold-pump experiment.}
\label{Table:MSE_tankManifold}
\begin{tabular}{|l|l|}
\hline
Manifold                    & \multicolumn{1}{c|}{\begin{tabular}[c]{@{}c@{}}Mean Square Error (MSE)\end{tabular}} \\ \hline
Tanks Height         & 9e-03                                                                                  \\ \hline
Volumetric Flow     & 2e-02                                                                                  \\ \hline
Area-Height         & 6e-03                                                                                  \\ \hline
Unseen inlet mass-flow & 1e-01                                                                                 \\ \hline
\end{tabular}
\end{table}

\subsection{Tank Network Modeling}
\subsubsection{Problem Description}
\begin{figure}[]
\centering
        \begin{overpic}[width=0.7\columnwidth]{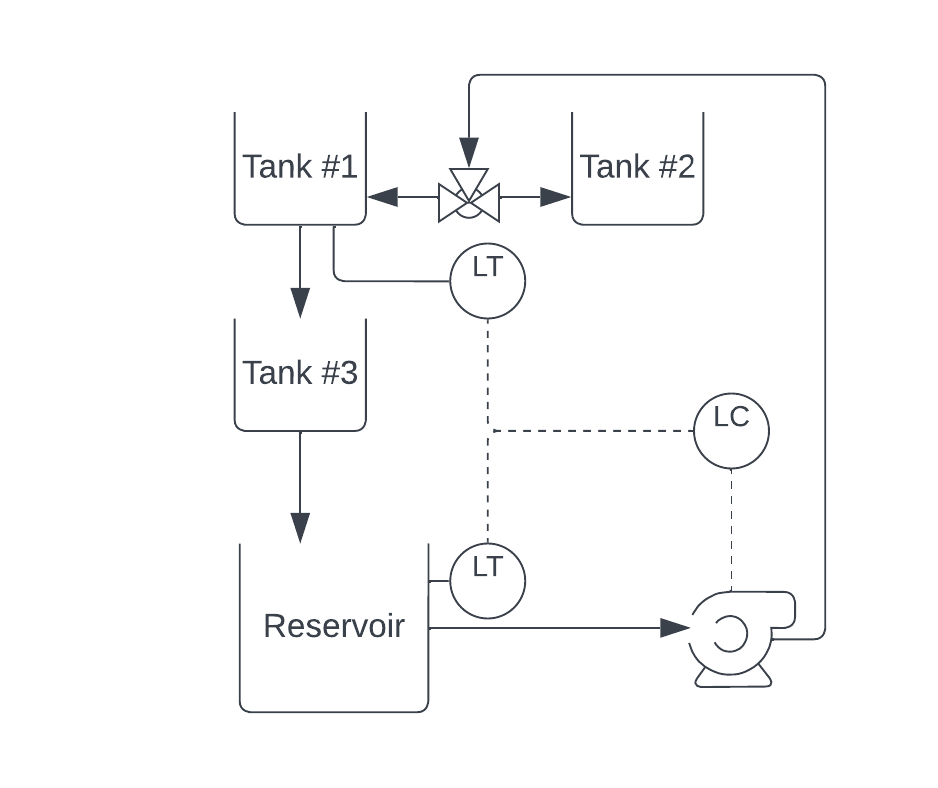}
        \end{overpic}  
        \caption{Networked system defined by tanks, pipes, a manifold, and a pump. The system exhibits oscillatory dynamics for some parameter choices and sensing schemes.}
        \label{fig:network}
\end{figure}

Figure \ref{fig:network} expands upon the tank-manifold problem of Fig. \ref{fig:manifold} by adding additional tanks, feedback mechanisms, and `closing the loop' in the network flow. In this example, the pump flow rate depends on the fluid heights of Tank \#1 and the reservoir. Tanks \#1 and \#2 have a height constraint (equivalent pressure head) and are fed by a common manifold. The outlets of tanks \#1 and \#3 depend on the square root of the fluid column height in the respective tanks. The full system is described by DAEs:
\begin{equation} \label{eq:network}
\begin{split}
    \frac{dx_1}{dt} &= \frac{1}{\phi_1(x_1)}\left(y_1 - y_3 \right), \quad
    \frac{dx_2}{dt} = \frac{1}{\phi_2(x_2)}\left(y_2 \right),\\
    \frac{dx_3}{dt} &= \frac{1}{\phi_3(x_3)}\left(y_3 - y_4 \right), \quad
    \frac{dx_4}{dt} = \frac{1}{\phi_4(x_4)}\left(y_4 - y_0 \right), \\
\end{split}
\end{equation}
\vspace{-8mm}
\begin{align}%
0 & = y_0 - p(x_1,x_4), \quad  0 = y_0 - y_1 - y_2,  \nonumber \\
0 & = y_3 - \alpha_1 \sqrt{x_1}, \quad  0 = y_4 - \alpha_2 \sqrt{x_3}, \ \quad 0 = x_1 - x_2.  \nonumber
\end{align}%
where $x$ denotes the fluid column height for tanks \#1 through \#4 (the reservoir), $p$ is a pump response function, $y$ represents volumetric flows, and $\alpha$ describe the discharge coefficients of the tank outlets. The differential states are $x = \{x_1,x_2,x_3,x_4\}\in\mathbb{R}^4$ and the algebraic states are $y=\{y_0,y_1,y_2,y_3,y_4\}\in\mathbb{R}^5$.
In this example, several characteristics make the dynamics more complex. First, the system is closed, as there is no material transfer into or out of the system. Second, there is the potential for backflow through the manifold from Tank \#2. Lastly, the feedback mechanism from the tank level transducers to the pump controller is such that oscillations can occur. 

\subsubsection{Optimization Problem}
The system given in Eq. \eqref{eq:network} was simulated with the following specifications:
\begin{align}%
\phi_1(x_1) = 2.0 \quad&  \phi_2(x_2) = 1.0  \nonumber \\
\phi_3(x_3) = 1.0 \quad&  \phi_4(x_4) = 10.0 \nonumber \\
p(x_1,x_4) = 0.1 x_1 x_4 \quad& \alpha_1, \alpha_2 = 0.1
\end{align}%

A single-trajectory dataset $\hat{X}$ was constructed through sampling this simulation at $\Delta t = 0.1$ for $t\in[0,20]$. The optimization problem definition is identical to that of Eq. \eqref{eq:manifold_loss}. For this problem, it is reasonable to assume that the basic physics for a gravity-fed pipe network is known, which provides a starting point for constructing a model. However, discharge coefficients and quantifying feedback mechanisms are not as straightforward. Here, we attempt to approximate (i) the Level Controller (LC) dynamics for pump control with a neural network $\NN_1:\mathbb{R}^2 \rightarrow \mathbb{R}^1$, (ii) manifold dynamics with with a second neural network $\NN_2:\mathbb{R}^2 \rightarrow \mathbb{R}^1$, and (iii) the discharge coefficients $\alpha_1$ and $\alpha_2$. The mappings $f$ and $h$ are defined as:
\begin{equation}
\begin{split}
        f : 
\begin{cases}
\frac{y_1 - y_3}{\phi_1 (x_1)}\\
\frac{y_2}{\phi_2 (x_2)}\\
\frac{y_3 - y_4}{\phi_3 (x_3)}\\
\frac{y_4 - y_0}{\phi_4 (x_4)}\\
\end{cases}, \quad
    h : 
\begin{cases}
    \NN_1(x_1,x_4;\theta_1) \\
    y_0\NN_2(x_2,x_3;\theta_2) \\
    y_0(1 - \NN_2(x_2,x_3;\theta_2)) \\
    \alpha_1 \sqrt{x_1} \\
    \alpha_2 \sqrt{x_3} \\
\end{cases}
\end{split}
\end{equation}

For this example, both $\NN_1$ and $\NN_2$ are simple multi-layer perceptrons, each with two hidden layers of size 30 and ReLU activations.

\subsubsection{Results}
\begin{figure}[]
    \centering
\includegraphics[scale=0.25,width=0.75\linewidth]{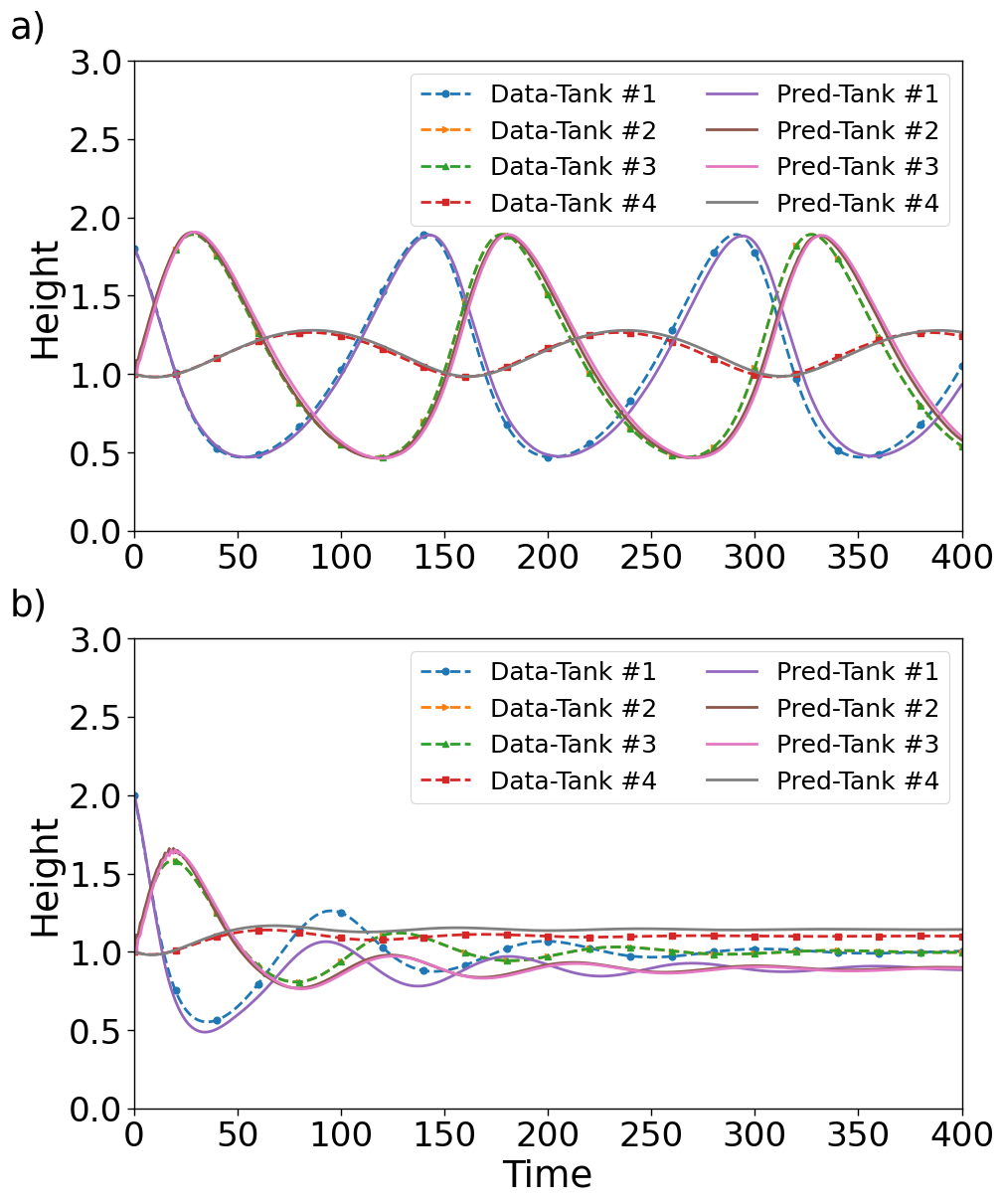}
    \caption{Tank Network problem: (a) Ground truth data vs NDAE model prediction. (b) Extrapolation case: inferring behavior for unseen component parameters using NDAE.}
    \label{fig:network_extrap_90}
\end{figure}

Figure \ref{fig:network_extrap_90}(a) shows open-loop trajectories
 of the four tank heights. At these process conditions, the pump delivers an oscillatory flow to the manifold, which our model is able to capture. Additionally, the tuned model can infer the values for the discharge coefficients: for the case of zero noise, the recovered values are $\alpha_1 = 0.1027$ and $\alpha_2 = 0.1024$. 
To evaluate generalization, we test the trained model on an unseen area–height relationship that was not used during training. Specifically, while the model is trained on one tank geometry with a particular cross-sectional area profile (training case), we deploy it on a different tank with a modified area–height profile (validation case); for this, we redefine $\phi_1(x_1) = 1.0 $. 
 In doing so, the system's limit cycle is destroyed, and the true dynamics approach a steady state. Figure \ref{fig:network_extrap_90}(b) demonstrates this extrapolation for the same time horizon. The reconstructed time series data matches the behavior of the unseen data: there exist damped oscillations that decay to a steady state value. Table-\ref{Table:TankNetwork_MSE} shows the MSE for the neural DAE models trained for the tank network problem. 

\begin{figure}[]
    \centering
\includegraphics[scale=0.25,width=0.75\linewidth]{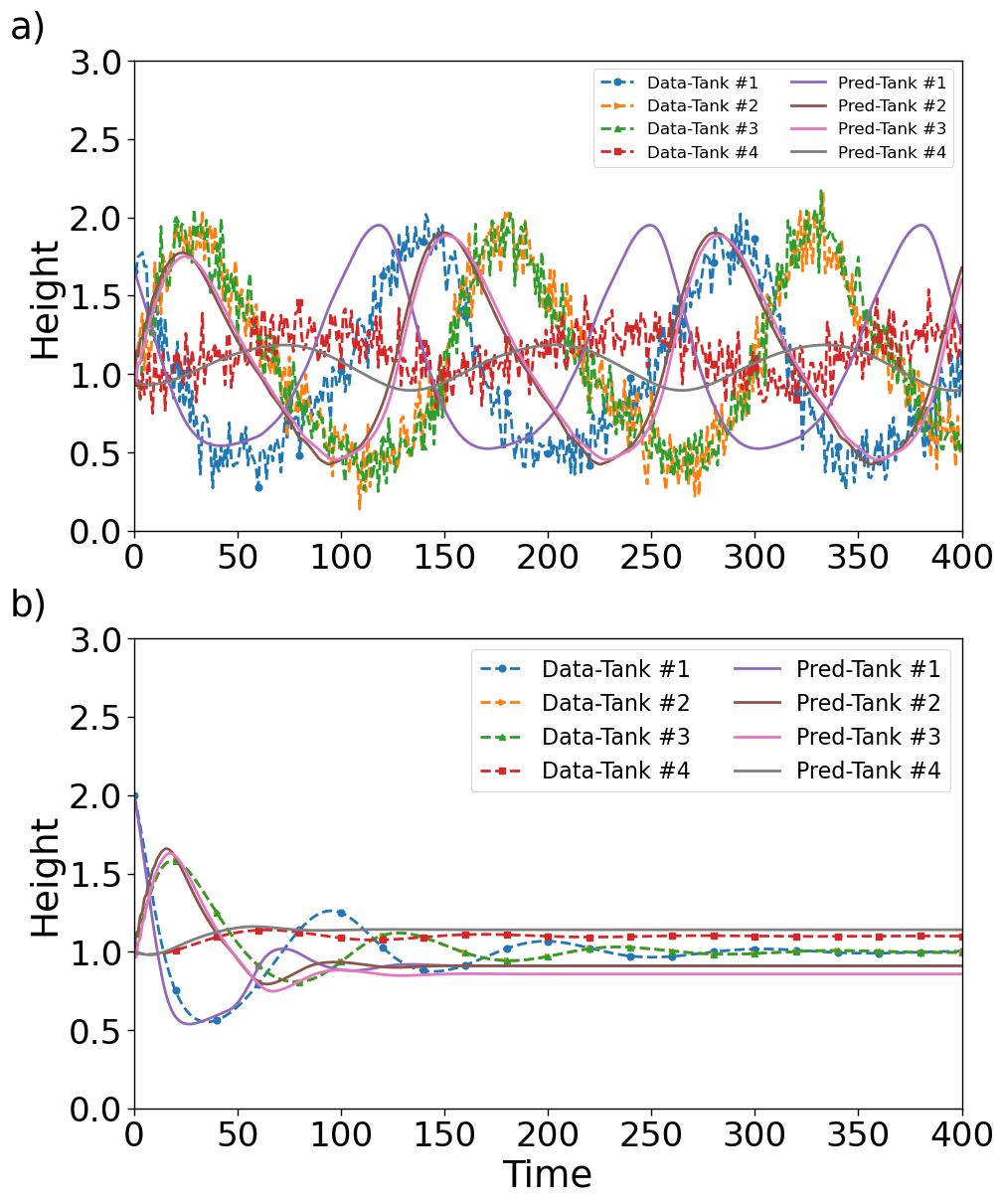}
    \caption{Tank Network problem: (a) noisy data vs Neural-DAE model prediction. (b) Extrapolation case: inferring behavior for unseen component parameters using Neural-DAE.}
    \label{fig:network_extrap_20}
\end{figure}

Noise tolerance is an important consideration in the evaluation of system identification tasks, especially in settings with nonlinearities and feedback mechanisms. For this example, we investigate noise tolerance by introducing Gaussian white noise. Figures \ref{fig:network_extrap_20}(a) and \ref{fig:network_extrap_20}(b) show similar training and extrapolation tasks for the Signal-to-Noise Ratio (SNR) of $20$ dB. Despite the significant noise, the learned dynamics are faithful to the underlying physics, as demonstrated in the extrapolation task, and result in an MSE of $7e-02$.

\begin{table}[h]
\centering
\caption{Mean Square error for the differential state and algebraic states for the tanks network experiment.}
\label{Table:TankNetwork_MSE}
\resizebox{\columnwidth}{!}{\begin{tabular}{|l|l|l|}
\hline
Trained Model Type                           & Tank-Network Experiment            & \multicolumn{1}{c|}{\begin{tabular}[c]{@{}c@{}}Mean Square Error\\ (MSE)\end{tabular}} \\ \hline
\multirow{2}{*}{No Noise}    & Tanks Height                       & 1e-02                                                                                  \\ \cline{2-3} 
                                            & Unseen initial Data : Tanks Height & 6e-02                                                                                  \\ \hline
\multirow{2}{*}{20 dB Noise} & Tanks Height                       & 1.03                                                                                   \\ \cline{2-3} 
                                            & Unseen initial Data : Tanks Height & 7e-2                                                                                   \\ \hline
\end{tabular}}
\end{table}

\section{CONCLUSIONS AND FUTURE WORKS} \label{sec:conclusion}

This work presents a data-driven framework for modeling differential algebraic equations (DAEs) using a novel operator splitting (OS) method for Neural DAEs.  Inspired by Lie-Trotter splitting, our approach sequentially updates algebraic states via a differentiable surrogate and integrates differential states using an ODE solver.
The effectiveness of the proposed method is demonstrated by (i) modeling networked dynamical systems with algebraic constraints and (ii) showing generalization to unseen process conditions or component variation. Furthermore, demonstrate robustness to noise and extrapolation to (i) learn the behaviors of the DAE system components and their interaction physics and (ii) disambiguate between data trends and mechanistic relationships contained in the DAE system.

Future research is needed in several areas. First, a baseline comparison with standard neural ODEs or PINNs would help quantify the gains of our approach. Second, a key assumption of the current formulation is that the dynamics are sufficiently smooth and non-stiff. To extend applicability to stiff or discontinuous dynamics, implicit integration schemes and adaptive solvers should be explored. Third, extending the framework to higher-index DAEs (characteristic of many optimal design and control applications) warrants further theoretical and numerical investigation. Lastly, this work has not addressed computational scalability or partial observability, both of which are critical for modeling large-scale systems with hundreds of degrees of freedom. We reserve these directions for future work.

\section*{ACKNOWLEDGMENTS}

This research was supported by the U.S. Department of Energy through the Building Technologies Office under the Advancing Market-Ready Building Energy Management by Cost-Effective Differentiable Predictive Control projects. PNNL is a multi-program national laboratory operated for the U.S. Department of Energy (DOE) by Battelle Memorial Institute under Contract No. DE-AC05-76RL0-1830.

This research was also supported by the Ralph O’Connor Sustainable Energy Institute at Johns Hopkins University.

\bibliographystyle{ieeetr}
\bibliography{references.bib}

\end{document}